\documentclass[11pt]{article}
\usepackage{acl2015}
\usepackage{mathptmx}
\usepackage[scaled=.9]{helvet}
\usepackage{courier}
\usepackage{mathtools}
\usepackage{dsfont}
\usepackage{amsmath,amssymb}

\usepackage{multicol, float}
\usepackage{fancyhdr}
\usepackage{geometry}
\usepackage{graphicx, subfigure}
\usepackage[normalem]{ulem}
\useunder{\uline}{\ul}{}
\usepackage{microtype}
\usepackage{algpseudocode}
\usepackage{algorithm}

\newcommand\ddfrac[2]{\frac{\displaystyle #1}{\displaystyle #2}}
\DeclareMathOperator*{\argmax}{arg\,max}

\title{Sequential Counterfactual Decision-Making Under Confounded Reward}

\author{Erik Skalnes \\
  {\tt eas2301@columbia.edu} \\}
\date{}

\usepackage[parfill]{parskip}

\begin{document}
\maketitle
\begin{abstract}

    We investigate the limitations of random trials when the cause of interest is confounded with the effect by formalizing a counterfactual policy-space where the agent's natural predilection is input to a soft-intervention. By connecting the counterfactual intervention in $G$ to a soft-intervention in a related graph called the conditional-twin, we prove graphical conditions under which the two policy types coincide in optimality. We also provide an estimation procedure for the distribution under the counterfactual-intervention, and connect it with the counterfactual distribution.  
    
\end{abstract}

\section{Introduction}

The canonical goal of a multi-armed bandit problem is to find $x \in X$ that maximizes reward $Y$. This means finding $x^* \in X$ such that 
$$ x^* = \argmax\limits_x E[Y \mid do(x)]$$

Algorithms like UCB or Thompson Sampling find $x^*$ by keeping an empirical distribution over reward for each arm. They explore arms with fewer samples and exploit arms with higher averages. If the empirical average over arms is about equal after many trials, it is tempting to conclude that the arms are about equal. In this paper, we give graphical conditions for when this type of experimentation may not correctly identify the optimal policy.

Past work has optimized \textit{the effect of the treatment on the treated} to differentiate empirically equal arms.$^1$ In MDP's, it was shown that algorithms that account for actions confounded with reward dominate algorithms that rely entirely on experimentation.$^2$ We prove conditions where such algorithms dominate experimental algorithms for any environment. We also formally prove why the algorithm in [1] works, and extend the results to the sequential-action case. 

The counterfactual randomization procedure described in this paper is relevant to any experimental pursuit where the environment is not fully known. Unobserved confounding can invalidate the conclusions of a random trial. We criticize experiments by proving their sub-optimality, and suggest a fix.  

\section{Background}

A fundamental goal of causal inference is to determine a layer 2 query, of the form $P(Y \mid do(x))$, from layer 1 data, of the form $P(V)$.$^3$ This is useful when interventions, like randomized control trials, cannot be performed. Even when they can, past work had to formalize the conditions under which the results of the trial can be extrapolated to new environments.$^4$ 

The primary contribution of this paper is to lay the theoretical foundations for estimating a layer 3 query from layer 2 data. The idea is that a decision confounded with reward allows inference about unobserved variables. This inference is only available in layer 1, because in layer 2 the confounding arrow is cut. 

We formalize the conditions under which such an inference is useful by presenting a generalization of a soft-intervention called the counterfactual intervention $\rho$. It accounts for the "natural predilection" of the decision-variable as it is determined by unobservable variables. We show that $\rho$ induces a sub-model on $G$ whose distribution is equivalent to the distribution induced by a soft-intervention $\pi$ in the conditional-twin of $G$. We then provide two examples of an estimation procedure in a realistic scenario, and use this procedure to formally explain the results in [1]. We also give graphical conditions for the advantage of the optimal counterfactual-intervention over the optimal soft-intervention based on the connection between value of information and d-separation.  

To ground the definitions, we consider the following causal graph implied by the SCM $M=<F, V, U, P(U)>$. 

\begin{figure}[H]
\centering
\includegraphics[width=0.8\linewidth]{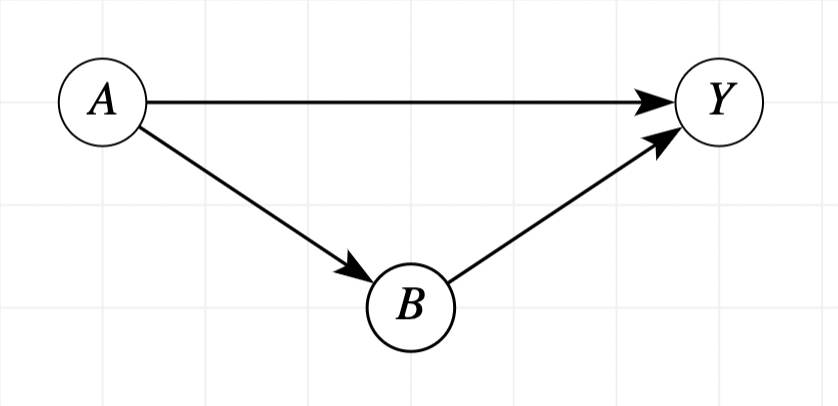}
\caption{Example Graph G}
\label{G}
\end{figure}

\noindent where \\
$A \leftarrow f_A(u_A)$ \\
$B \leftarrow f_B(a, u_B)$ \\
$Y \leftarrow f_Y(a, b, u_Y)$ \\
and $P(U)$ is arbitrary

We refer to the twin graph of $G$ as $G'$ and note that $Y$ is a reward variable. That is, the objective of the intervention is to estimate the distribution over $Y$.

\section{Counterfactual Interventions}

Let $N \in V$ be an observable node with natural value $f_N(pa(N), U_N)$. Let $\rho_N$ be a \textbf{counterfactual intervention} if 
$\rho_N: N \rightarrow N$ maps $f_N(pa(N), U_N) \rightarrow P(n \in N)$. That is, $\rho_N$ maps the natural value of $N$ under a setting of $U$ to a distribution over the domain of $N$.

The key difference from a soft intervention $\pi_N: A \in An(N) \rightarrow P(n\in N)$ is that $\rho$ maps information that depends on $f_N$ and $U_N$, whereas $\pi_N$ can only account for $An(N)$.

We can extend the definition to more than one variable by defining a counterfactual intervention on a set of variables as a set of counterfactual interventions on each variable in the set. Let $I \subseteq V$ be a set of "intuition" variables and $\rho_I$ be a set of counterfactual interventions 
$\rho_I := \{ \rho_i \mid \rho_I: f_I(U) \rightarrow I \}$
then $\rho_I$ is a set of \textbf{counterfactual interventions}.

Consider $\rho_I$ for $I = \{A, B\}$ on $G$

\begin{figure}[H]
\centering
\includegraphics[width=.8\linewidth]{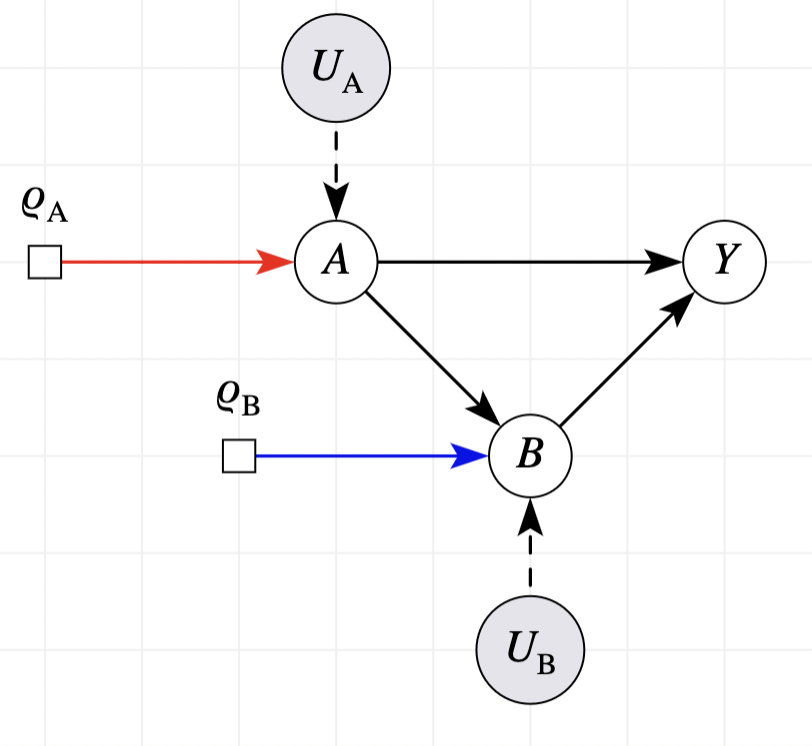}
\caption{$G_{\rho_I}$}
\label{G_rho}
\end{figure}

\noindent which induces the functional form \\
$A \leftarrow \rho_A(f_A(u_A))$ \\
$B \leftarrow \rho_B(f_B(\rho_A, u_B))$\\
$Y \leftarrow f_Y(\rho_A, \rho_B, u_Y)$ 

Note two salient features: (1) the counterfactual intervention does not cut incoming arrows like atomic or soft interventions, and (2) the natural value $f_B$ that $\rho_B$ changes is dependent on $\rho_A$. Fact (2) motivates the definition of a conditional twin graph. 

\section{Conditional Twin}

Before motivating the conditional-twin, it is important to motivate a soft-intervention on the twin graph as it relates to the counterfactual intervention. Then we prove a theorem equivocating the distribution in $G$ under a counterfactual-intervention with the distribution in the conditional-twin of $G$ under a soft-intervention. 

\subsection{Motivation}

 In the single-action case with action $A$, the counterfactual intervention on $G$ is equivalent to a soft-intervention on $G'$. Consider $G_{\rho_A}$, the sub-model induced by $\rho_I = \{ \rho_A\}$ on $G$:

$A \leftarrow \rho_A(f_A(U_A))$ \\
$B \leftarrow f_B(A, U_B)$ \\
$Y \leftarrow f_Y(A, B, U_Y)$

The value of $A$ that is input to $B$ and $Y$ is $\rho_A(f_A(U_A))$. The value of $B$ that is input to $Y$ is $f_B(\rho_A(f_A(U_A)), U_B)$. No variables after $A$ in the topological order 'see' $f_A$.

Consider the sub-model induced by $\pi_A: a\in A \rightarrow P(a \in A')$ on $G'$, a soft-intervention on the twin graph of $G$. 

\begin{figure}[H]
\centering
\includegraphics[width=.8\linewidth]{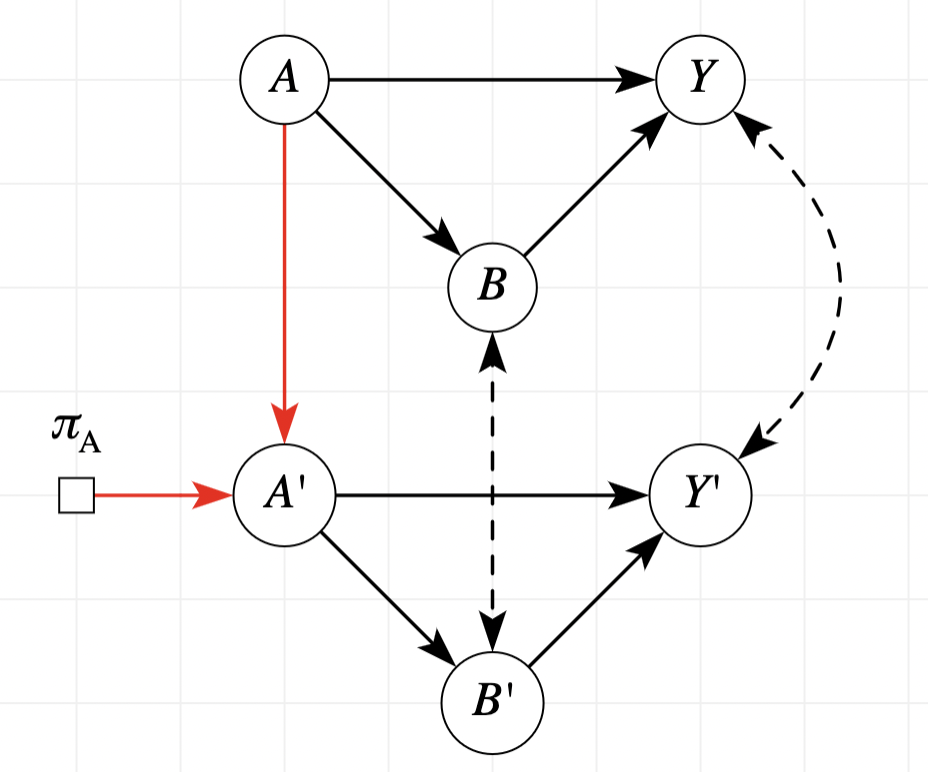}
\caption{$G'_{\{\pi_A\}}$}
\label{G_rho}
\end{figure}

$A \leftarrow f_A(U_A)$ \\
$B \leftarrow f_B(A, U_B)$ \\
$Y \leftarrow f_Y(A, B, U_Y)$ 

$A' \leftarrow \pi_A(f_A(U_A))$ \\ 
$B' \leftarrow f_B(A', U_B)$ \\
$Y' \leftarrow f_Y(A', B', U_Y)$

Let's call $A,B,Y$ row0. The distribution over variables in row0 is the same as the distribution over variables in $G$ with no intervention. Let's call $A', B', Y'$ row1. The distribution over variables in row1 is the same as the distribution over variables in $G_{\rho_A}$ when $\pi_A = \rho_A$. This is simply because their sub-models are equivalent. The sense in which $G_{\rho_A}$ and $G'_{\pi_A}$ are equivalent is that any distribution induced in $G$ is also induced in one of the rows of $G'$. 

The motivation for the conditional twin comes from trying to simulate $\rho_I$ for $I=\{A, B\}$. The sub-model induced by $\rho_I$ on $G$ is the graph in figure 2: 

$A \leftarrow \rho_A(f_A(U_A))$ \\
$B \leftarrow \rho_B(f_B(\rho_A(f_A(U_A)), U_B))$ \\
$Y \leftarrow f_Y(A, B, U_Y)$

In $G'$, it is impossible to design a soft intervention that both sets $A \leftarrow \rho_A$ and $B \leftarrow \rho_B$. If it were possible, it would have to include $\pi_A$ to make $A' \leftarrow \pi_A(f_A(U_A))$. Now, $\pi_B$ must take into account the value of $f_B$. Any soft-intervention on $B'$ will cut the arrow from $A'$ to $B'$, making it impossible to get $f_B(A')$. The only way to simulate $G_{\rho_I}$ in $G'_{\pi_A}$ is to perform a $\textit{counterfactual intervention}$ $\rho_B$ on $B'$. 

We already know that we can simulate a single counterfactual intervention with a soft-intervention on the twin graph. Since $\rho_B$ inputs $A'$ and by necessity of $\rho_B$ accounting for $\pi_A$ which accounts for $f_A(U_A))$, we create the twin with respect to the sub-graph $G'_{B'Y'}$. Conceptually, the direct effect of $A'$ on $Y'$ has been decided. Only the value of $B'$ is being intervened on. Thus, we can simulate the distribution of $Y'$ in $G'_{\pi_A, \rho_B}$ with a soft-intervention on $G''_{B'Y'}$, the twin graph of the sub-graph of $G'$ containing only the nodes $B', Y'$. 

\begin{figure}[H]
\centering
\includegraphics[width=.8\linewidth]{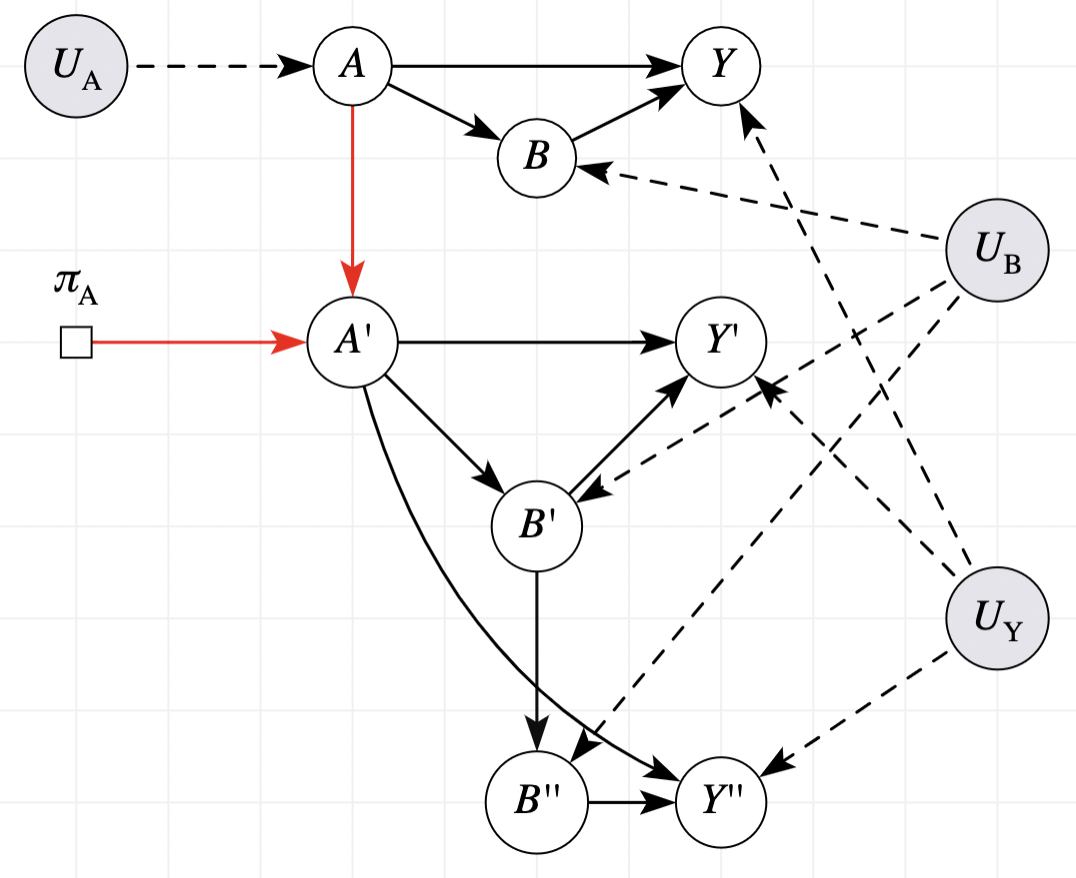}
\caption{$G'' = twin(G'_{B'Y'} \mid A' = \pi_A)$}
\label{G_rho}
\end{figure}

Consider the soft-intervention $\pi_B = \rho_B$ on this graph

\begin{figure}[H]
\centering
\includegraphics[width=.8\linewidth]{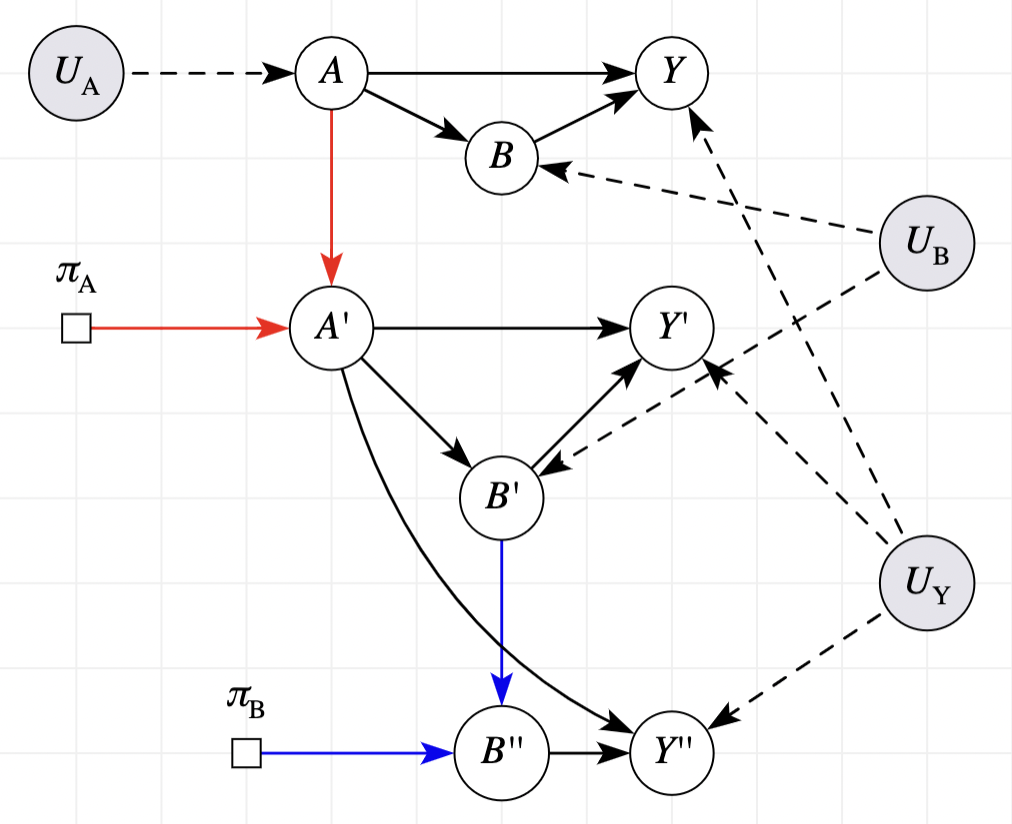}
\caption{$G''_{\pi_A \pi_B}$}
\label{G_rho}
\end{figure}

Now $\pi_B$ gets to see the natural value of $B$ conditional on $\pi_A$, which was impossible before. Indeed, the distribution over $Y''$ in $G''$ is the same as the distribution over $Y_{\rho_I}$ for $\rho_I = \pi_I$. 

$Y'' \leftarrow f_Y(\pi_A(A), \pi_B(f_B(\pi_A, U_B)), U_Y)$ \\
is equivalent to \\
$Y_{\rho_I} \leftarrow f_Y(\rho_A(A), \rho_B(f_B(\rho_A, U_B)), U_Y)$ 

It should be further noted that neither $Y$ not $Y'$ simulate any distribution that cannot be simulated with $Y''$. So we could simulate $Y_{\rho_{AB}}$ with the same $G''$ but having removed the nodes $B, Y, Y'$. 

What remains is to formalize the construction of the conditional twin, so that we can prove that the analog of $Y''$ for the general case can simulate $Y_{\rho_I}$ for any $I$ on any SCM. 

\subsection{Formal Construction}

A graph $G^c$ is the $\textbf{conditional twin}$ of input $G,I$ \\ 
$\iff$ it is the output of algorithm 1. 

Input: $G$ and $I = \{I_1 ,..., I_k \} \subseteq (V \setminus Y)$ s.t. \\
$i < j \implies I_i \notin An(I_j)$ (Topological Order)

Output: a graph $G^c$ and a soft-intervention 
$$\pi_I = \{\pi_{I'_j}: i \in I_j \rightarrow \rho_{I_j}(i) \text{ for } I'_j \in I'^c\}$$
such that 
$P_{G_{\rho_I}}(Y) = P_{G^c_{\pi_I}}(Y)$

In words, input a graph $G$ and a set of variables $I$ on which some counterfactual-intervention $\rho_I$ can take place. We do not specify the functional form of $\rho_I$. Output a conditional-twin graph $G^c$ and a soft-intervention $\pi_I$. The sub-model $G^c_{\pi_I}$ has the same distribution over variables as the sub-model $G_{\rho_I}$, regardless of the functional form of $\rho_I$.  

\begin{algorithm}
\caption{ConditionalTwin(G, I)}\label{cond_twin}

1. \textbf{Require} $I = I_1 ,\ldots, I_k$ in topological order \\
2. \textbf{if} {$I = \emptyset$}: return G  \\
3. create node $I_1'$ in G \\
4. $I_1 \xrightarrow{i_1}$ becomes $I_1 \xrightarrow{\pi_{I_1}(i_1)} I_1' \xrightarrow{i_1'} $ \\
5. ConditionalTwin(G, $I = I_2, \ldots, I_k$)
\end{algorithm} 

explanation: \\
(1) input $I$ in topological order w.r.t $G$. \\
(2) handle all intuition variables. \\
(3) add a node $I_1'$ to $G$. \\
(4) In the functional forms of children of $I_1$, replace $I_1$ with $I_1'$. $I_1$ retains its parents, its only child is $I_1'$, and $I_1' \gets \pi_{I_1}(i_1 \in I_1)$ in $F$. Once $\rho_I$ is defined, $\pi_{I_1}$ goes from the identity function to $\pi_{I_1} = \rho_{I_1}$. \\ 
(5) recurse on $I$ with $I_1$ removed. In the jth iteration iteration, $I_j$ will be $I_1$. 

The conditional twin returned on our example $G$ is the following:

\begin{figure}[H]
\centering
\includegraphics[width=.8\linewidth]{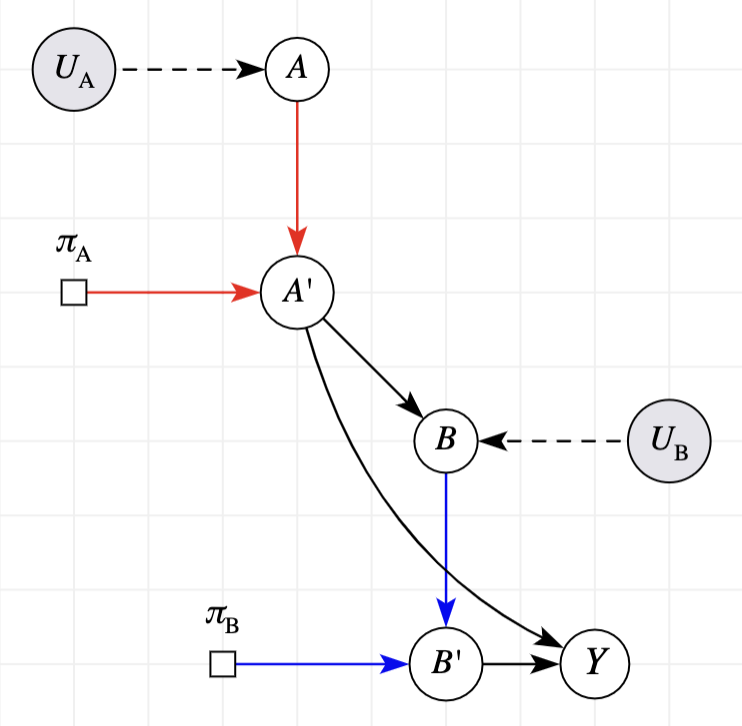}
\caption{$G^c = \text{ConditionalTwin}(G, I=\{A, B\}$)}
\label{G_rho}
\end{figure}

$G^c$ is similar to the graph $G''$ in figure 5 in that $P_{G''}(Y'') = P_{G^c}(Y)$. This is because they have the same inputs in $F$. ConditionalTwin() does not include the superfluous variables $B,Y, Y'$, whose distributions can be simulated by other variables in $G^c$. For example, simulate $Y'_{G''}$ with $Y_{G^c}$ by leaving $\pi_B(b) = b$ and letting $\pi_A = \rho_A$. 

In the next section, we show that Algorithm1 generalizes to any $G, I$. A conditional twin and corresponding $\pi$ can always simulate the counterfactual intervention. 

\subsection{Theorem 1: Counterfactual-Interventional Equivalence }

Let $M: = <V, U, F, P(U)>$ be an arbitrary SCM with corresponding $G$.

$I  := \{I_1 ,..., I_k \} \subseteq V \setminus Y$ \\
$\rho_I := \{  \rho_{I_j}: I_j \rightarrow I_j \}$ \\
$G^c := \text{ConditionalTwin}(G, I)$ \\
$\pi_I := \{ \pi_{I_j}: I_j \rightarrow I_j \mid \pi_{I_j}(i_j) = \rho_{I_j}(i_j)\}$ 

Denote a distribution in $G$ under $\rho_I$ as $P_\rho(\dot)$ and a distribution in $G^c$ under $\pi$ as $P_\pi()$. Denote a variable $N$ in $G$ as $N$ and a variable $N$ in $G^c$ as $N^c$. The 'counterfactual copy' of a variable $I_j \in I$  that exist in $G^c$ only, created by step 3 of algorithm 1, is referred to as $I_j'$. 
\textbf{Lemma1} \\
$P_\rho(I) = P_\pi(I')$ for any $G,\rho_I$

\textbf{Proof} 

\underline{Base Case}

let $I = \{I_1 \}$

Lemma1 is implied if $I_1$ has the same inputs as $I'_1$ because if they have the same inputs, they take the same value under any setting of $U$.  

$I_1 \gets \rho_{I_1}(f_{I_1}(Pa(I_1), U_{I_1}))$ by definition of $\rho$

$I_1^c \gets f_{I_1}(Pa(I_1), U_{I_1})$ \\
$I'_1 \gets \pi_{I_1}(I_1)$ \\
by construction of $G^c$

$\pi_{I_1} = \rho_{I_1}$ by definition of $\pi$ \\ 
thus 
$I_1' \gets \rho_{I_1}(f_{I_1}(Pa(I_1), U_{I_1})$

Since their functional inputs are the same, the distributions are the same for any $U$. 

\underline{induction}

let $k \geq 1$ \\
$I = \{ I_1 ,\ldots, I_k \} \; $ \\
$J = \{ I_1 ,\ldots, I_k, I_j\}$ \\
$I_j \in V$ and $I_j  \notin An(I)$ 

Prove: \\
$P_{\rho_I}(I) = P_{\pi_I}(I') \implies 
P_{\rho_J}(J) = P_{\pi_J}(J')$

let $GI^c$ = ConditionalTwin(G, I) \\
let $GJ^c$ = ConditionalTwin(G, J)

$GJ^c = \text{ConditionalTwin}(GI^c, \{I_j\})$ \\
because $I_j$ is the last element of topologically sorted $J$. 

Consider the base case, where $G=GI^c, I=\{I_j\}$. Since $I_j \notin An(I)$, the intervention on $I_j$ cannot affect the distribution over $I$. Thus, $P_{\rho_J}(J) = P_{\pi_J}(J')$. 

The distributions are equal for the first step of algorithm1 on arbitrary $G, I$. It is a recursive algorithm, so the distributions remain equal for each step. 
\textbf{Theorem} \\
$P_{\rho}(V) = P_{\pi}(V^c \setminus I_1)$

\textbf{Proof}

For an arbitrary $I_j \in I$, \\
$P_\rho(I_j) = P_{\pi}(I'_j)$ by lemma1 \\
$De(I_j) = De(I'_j)$ by construction of $G^c$. \\
Thus, $P_\rho(De(I_j)) = P_{\pi}(De(I'_j))$

Ancestors of $I$ and variables not connected to $I$ on any undirected path are not affected by the intervention. By construction of $G^c$, they take the same values in $G^c_\pi$ as they do in $G_\rho$. 

\section{Estimating A Counterfactual}

We can calculate distributions under $\rho$ if we have the conditional twin. We argue that realistic scenarios described by $G$ exist where $G^c$ can be induced. We further show that a counterfactual distribution can be estimated by the counterfactual intervention in the single-action case. We formulate but leave open the question of whether a counterfactual distribution can be estimated in the multi-action case.   

\subsection{Inducing the Conditional Twin}

Doc is driving a car. He can turn the wheel left or right, represented by variable $A \in \{0,1\}$. He gets reward $Y$ for not crashing. Both $A$ and $Y$ are affected by environmental conditions $U_{AY}$. 

\begin{figure}[H]
\centering
\includegraphics[width=.7\linewidth]{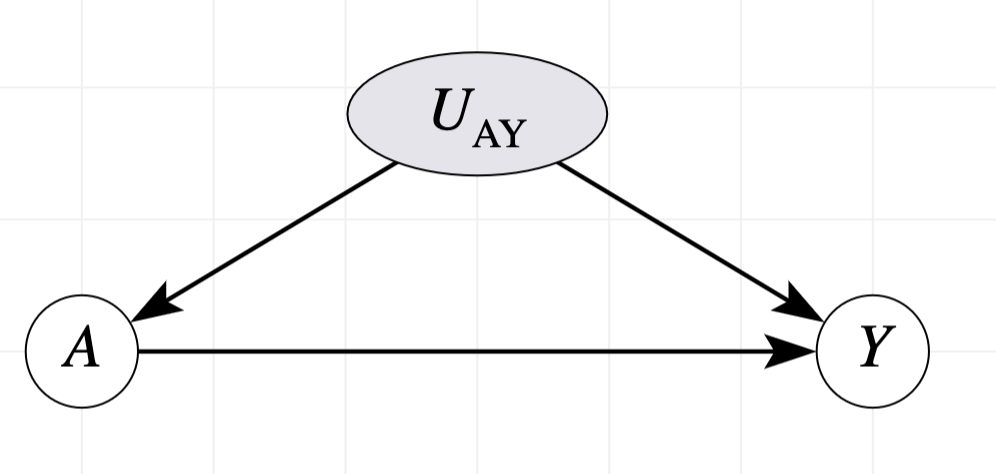}
\caption{$H$: Driving with Doc}
\label{G_rho}
\end{figure}

Rob is a computer in the wheel who senses Doc's choice and has the final say of which direction the car goes. Sometimes he changes it, and sometimes he doesn't. Rob's choice is represented by the node $A'$, and his decision-making algorithm is represented by $\pi_A$.  

\begin{figure}[H]
\centering
\includegraphics[width=.7\linewidth]{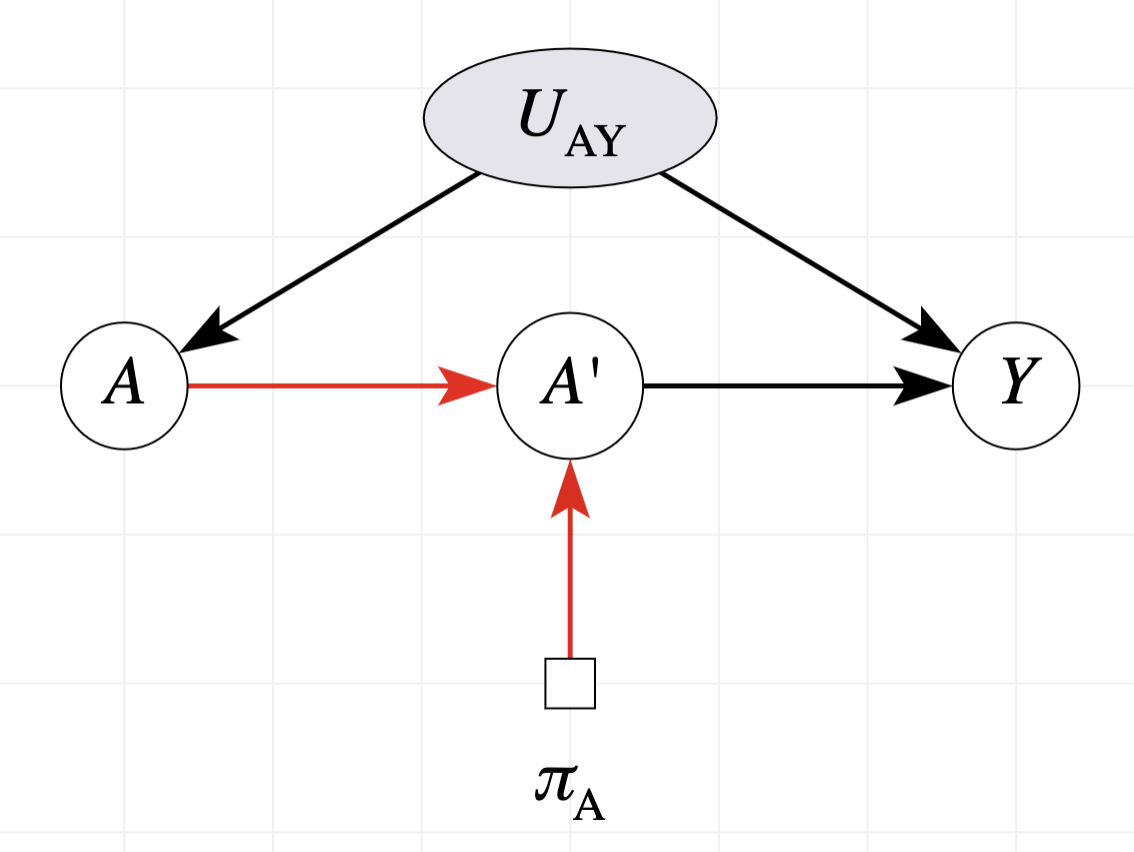}
\caption{$H^c$: Driving with Doc and Rob}
\label{G_rho}
\end{figure}

$H^c$ is the conditional twin of $H$. The assumptions are that road conditions $U$ do not change in the time it takes to intervene on Doc's decision, and that Doc's decision can be accurately sensed. In the single-action case represented by graph $H$, we can show that estimating $P(Y_{\rho_A})$ is the same as estimating $P(Y_{a'} \mid a)$.  

\subsection{Single-Action}

We can estimate $P(Y_{A'=\pi_A}=1 \mid A=a)$ in the standard way. 
$$P(Y_\pi = 1)$$
$$= \ddfrac{\text{frequency with which Y=1 under } \pi}{\text{number of trials}}$$

thus,
$$ \hat P_H(Y_{\rho_A} = 1)$$
$$ = \hat P_{H^c}(Y_{A' = \pi_A} =1 \mid A=a)$$
\begin{equation} 
= \ddfrac{1_{A=a, A' = \pi_A, Y=1}}{1_{A=a, A' = \pi_a}}
\end{equation}
$$ \approx \sum\limits_{u \mid A(u) = a, Y_{a'}(u) = 1} P(u)$$
$$ = P(Y_{a'} =1 \mid a)$$

Which is the counterfactual. The counterfactual is the natural regime when $A' = A$. By equation (1),
$$ P(Y_{A \neq a} =1 \mid A = a)$$
$$ \approx \ddfrac{1_{A=a, A' \neq a, Y=1}}{1_{A=a, A' \neq a}}$$

We can estimate a counterfactual probability of $Y=1$ in $H$ by estimating the frequency with which switching the value of $A$ in $H^c$ caused $Y=1$. This is because the soft-intervention in $H^c$ is equivalent to the counterfactual-intervention in $H$, and the counterfactual-intervention in $H$ is equivalent to the counterfactual distribution in $H$. The last part is open is the multi-action case. 

\subsection{Multi-Action}

Consider a similar story in $G$. Let's say that $A$ is the turn-signal and $B$ is the wheel. Rob can intervene with $\pi = \{ \pi_A, \pi_B \}$. 

Consider the familiar counterfactual
\begin{equation}
P(Y_{A\neq a, B \neq b} = 1 \mid a, b)
\end{equation}
$$= \sum\limits_{u \mid A =a, B=b, Y_{a'b'}(u) = 1} P(u)$$
$$ \neq \ddfrac{1_{A=a, A' \neq a, B=b, B' \neq b,  Y=1}}{1_{A=a, A' \neq a, B=b, B' \neq b}}$$
$$ = \hat P_{H^c}(Y_{A' \neq a, B' \neq b} =1 \mid A=a, B=b)$$

The issue is that we can only observe $B$ after $A$ has a value. Formally,

Let $u_{ab} \in U_{ab} \subseteq U$ be an arbitrary element in the domain of $U$ that induces in $G$ \\
$f_A(u_{ab}) = a$ \\
$f_B(u_{ab}, b) = b$

The observation of $A=a, B=b$ in the natural regime is proof that the setting of $u$ is a member of $U_{ab}$. Under $\rho_I$ however, the observation of $B$ depends on the value of $\rho_A$. Thus, $A = a, B=b$ is not evidence that $u \in U_{ab}$, unless $\rho_A(a)$ is always equal to $a$. This is because $A$ is an ancestor of $B$, so we can only observe $B$ conditional on the value of $\rho_A$. 

However, we are not without recourse. Consider the derivation of the probability of $Y_\rho$:

$$P(Y_{\rho_{I = \{AB\}}} = 1)$$
$$= \sum\limits_{a} P(Y_{\rho_I}= 1 \mid \ A = a)P(A=a)$$
$$= \sum\limits_{a} P(Y_{A \neq a, \rho_{B}} = y \mid \ A = a)P(A=a)$$
$$= \sum\limits_{ab} P(Y_{A \neq a, B \neq b} = 1 \mid A = a, B_{a'}=b)$$ $$P(B_{a'} = b \mid A=a) P(A=a)$$
$$= 
\bigg[ \sum\limits_{u_{ab}} P(u) \bigg]
\bigg[ \sum\limits_{u_b} P(u) \bigg]
\bigg[ \sum\limits_{u_a} P(u) \bigg]$$

where 

$$ u_{ab} \mid A(u)=a \wedge B_{a'}(u)=b \wedge Y_{a'b'}(u_3) = 1 $$
$$ u_b \mid A(u) = a \wedge B_{a'}(u) = b$$ 
$$ u_a \mid A(u) = a $$

Each of the three pieces can be estimated.
$$ P(Y_{A \neq a, B \neq b} = 1 \mid A = a, B_{a'}=b)$$
$$= \ddfrac{1_{A=a, A' \neq a, B=b, B' \neq b,  Y=1}}{1_{A=a, A' \neq a, B=b, B' \neq b}}$$
$$ P(B_{a'} = b \mid A=a) = \ddfrac{\mathds{1} \{ A=a, A\neq a, B=b \}}{\mathds{1} \{ A=a, A\neq a \}}$$
$$P(A=a) = \ddfrac{\mathds{1} \{ A=a\}}{\mathds{1} \{ \text{True}\}}$$

These are estimators of the counterfactual intervention. Future work could determine how to connect them to a counterfactual distribution.

\section{Optimality Conditions for  $\rho$}

The computational costs associated with computing $\rho$ are strictly higher than computing $\pi$ because the search space is bigger. For each observation, $\rho$ decides on an intervention, whereas $\pi$ is merely an intervention. It is therefore desirable to have graphical conditions for when $\rho$ could be superior. 

Consider an arbitrary graph $G$ and a variable $N$. We will discuss three types of interventions: \\
$\rho$ denotes a counterfactual intervention in $G$\\
$\pi$ denotes the corresponding soft-intervention in $G^c$\\
$\sigma$ denotes a soft-intervention in $G$. 

In general, $E[Y_\rho] \geq E[Y_\sigma]$ because $\rho$ has access to more information than $\sigma$. 
We want to determine when $E[Y_\rho] > E[Y_\sigma]$ because we want to know when the counterfactual intervention is better than an experiment. 

Let 
$$\rho:f_N \rightarrow P(N)$$ 
$$\rho^* = \argmax\limits_\rho E_{G_\rho}[Y]$$ 
$$\pi: N \rightarrow P(N') \text{ in } G^c \text{ mimic }  \rho$$ 
$$\pi^* = \argmax\limits_\rho E_{G^c_\pi}[Y]$$
Theorem 1 implies \textbf{Corollary 1}
$$E_{G_{\rho^*}}[Y] = E_{G_{\pi^*}^c}[Y]$$

Let 
$$\sigma: \emptyset \rightarrow P(n \in N) \text{ in } G $$ 
$$ \sigma^* = \argmax\limits_\sigma E_{G_\sigma}[Y]$$
$$\rho_{\emptyset}: \emptyset \rightarrow P(N)$$
$$\pi_{\emptyset}: \emptyset \rightarrow P(N') \text{ in } G^c \text{ mimic }  \rho_\emptyset$$

The counterfactual intervention $\rho_{\emptyset}$ completely ignores $f_N$. It inputs $f_N$ but does not use it to construct $P(N)$. The will be no arrow between $I$ and $I'$ in the conditional-twin under $\pi_\emptyset$. 

Theorem 1 implies \textbf{Corollary 2}
$$E_{G_{\rho^*_{\emptyset}}}[Y] = E_{G^c_{\pi^*_\emptyset}}[Y]$$

By the Corollaries,
$$ E_{G_{\rho^*}}[Y] = E_{G^c_{\pi^*}}[Y]$$
$$ E_{G_{\sigma^*}}[Y] = E_{G_{\rho_\emptyset^*}}[Y] =
E_{G^c_{\pi^*_\emptyset}}[Y]$$
Thus
$$ E_{G_{\rho^*}}[Y] > E_{G_{\sigma^*}}[Y] \iff$$
\begin{equation}
E_{G^c_{\pi^*}}[Y] > E_{G^c_{\pi^*_\emptyset}}[Y]
\end{equation}

In words, the optimal counterfactual intervention is better than the optimal experiment if and only if the optimal $\pi$ in the conditional-twin is better than the optimal experiment in the conditional-twin. 

$\pi$ inputs strictly more information than $\pi_\emptyset$ because $\pi$ inputs $N$. Thus, (3) holds when the $\textit{increase in E[Y] of observing }N > 0$. 

$\textbf{Definition: Value of Observation (VO)}$\\
Let $\pi_I$ be a soft-intervention on a set of variables $I$ in a causal graph $G$. 
$$ \text{VO}(N) := E[Y \ ;\ \pi_I] -  E[Y \ ;\ \pi_{I \setminus N}]$$

Thus d-separation is a sufficient condition for $VO(N)$ to be zero. 
$$ P(Y_\pi \mid N) = P(Y_\pi) \implies VO(N) = 0$$

Note that the reverse is not true: the value of a $N$ may be zero and be d-connected with $Y$. For example, when $N$ points to $Y$ but $f_Y = 1$. Thus, this condition only tells us when the counterfactual intervention $\textit{could}$ be better than an experiment. 

This result extends to the multi-action case because of the recursive construction of the conditional twin. If the $VO$ of a single variable is not 0, then the counterfactual intervention will be better than the empty-soft-intervention. Only when the $VO$ of all variables is 0 does (3) have equality. 

\section{Conclusion and Future Work}

We define $\rho$ and give conditions under which it could be superior to $\sigma$. The point is that an agent that employs an experimental procedure cannot be sure that they incur sub-linear regret, even after many samples, if the conditions hold. These environments require a different policy-space that accounts for the natural state of the system, which we call the counterfactual-intervention $\rho$. In the sequential-action case, $\rho$ can be understood as a soft-intervention on a graph called the condition-twin. This observation provides the theoretical basis for estimating counterfactuals using layer 2 data.

Future work should try to connect $\rho$ with counterfactual distributions. If that can be done, then we will have a pipeline for computing counterfactuals with layer 2: $\pi$ on $G^c \rightarrow \rho$ on $G \rightarrow$? counterfactual distribution. A good starting place would be to prove that any single-intervention counterfactual distribution can be estimated in the conditional-twin. Extensions of the $\rho$ policy-space to account for ancestors would also be a good idea.


\begin{thebibliography}{}

\bibitem[\protect\citename{Elias Bareinboim and Judea Pearl}2015]{Bareinboim:15}
[1] Elias Bareinboim and Judea Pearl
\newblock 2015.
\newblock {\em Bandits with Unobserved Confounders.}
\newblock Advances in Neural Information Processing Systems 28 (NIPS 2015). 

\bibitem[\protect\citename{Junzhe Zhang and Elias Bareinboim}2022]{Zhang:22}
[2] Junzhe Zhang and Elias Bareinboim
\newblock 2022.
\newblock {\em Can Humans be out of the Loop?}
\newblock CLeaR 2022.  

\bibitem[\protect\citename{ Judea Pearl}2009]{Pearl:09}
[3] Judea Pearl
\newblock 2009.
\newblock {\em Causality: Models, Reasoning, and Inference}
\newblock (Cambridge Univ Press, New York), 2nd Ed

\bibitem[\protect\citename{Elias Bareinboim and Judea Pearl}2015]{Bareinboim:15}
[4] Elias Bareinboim and Judea Pearl
\newblock 2015.
\newblock {\em Causal Inference and the Data-Fusion Problem}
\newblock Proceedings of the National Academy of Sciences 2015

\bibitem[\protect\citename{Juan Correa and Elias Bareinboim}2020]{Correa:20}
[5] Juan Correa and Elias Bareinboim
\newblock 2020.
\newblock {\em A Calculus for Stochastic Interventions:
Causal Effect Identification and Surrogate Experiments}
\newblock Association for the Advancement of Artificial
Intelligence 



\end{thebibliography}
\end{document}